\documentclass[sigconf,screen]{acmart}

\usepackage{graphicx}
\usepackage{amsmath}
\usepackage{multirow}
\usepackage{booktabs}
\usepackage{diagbox}

\usepackage{algorithm}
\usepackage[noend]{algpseudocode}
\usepackage{xspace}
\usepackage{xcolor}
\usepackage[misc]{ifsym}

\makeatletter
\DeclareRobustCommand\onedot{\futurelet\@let@token\@onedot}
\def\@onedot{\ifx\@let@token.\else.\null\fi\xspace}
\def\eg{\emph{e.g}\onedot} 
\def\ie{\emph{i.e}\onedot} 
 
\def\etc{\emph{etc}\onedot}

\makeatother

\graphicspath{ {figures/} }

\AtBeginDocument{%
  \providecommand\BibTeX{{%
    \normalfont B\kern-0.5em{\scshape i\kern-0.25em b}\kern-0.8em\TeX}}}

\copyrightyear{2022}
\acmYear{2022}
\setcopyright{acmcopyright}\acmConference[MM '22]{Proceedings of the 30th ACM International Conference on Multimedia}{October 10--14, 2022}{Lisboa, Portugal}
\acmBooktitle{Proceedings of the 30th ACM International Conference on Multimedia (MM '22), October 10--14, 2022, Lisboa, Portugal}
\acmPrice{15.00}
\acmDOI{10.1145/3503161.3547936}
\acmISBN{978-1-4503-9203-7/22/10}

\begin{document}
%
\title{Towards Unbiased Visual Emotion Recognition via Causal Intervention}

\author{Yuedong Chen} 
\authornote{\Letter~Corresponding Author}
\email{yuedong.chen@monash.edu}
\orcid{0000-0003-0943-1512}
\affiliation{%
  \institution{Monash University}
  \city{Melbourne}
  \country{Australia}
}

\author{Xu Yang}
\email{101013120@seu.edu.cn}
\orcid{0000-0002-8276-2679}
\affiliation{%
  \institution{Southeast University}
  \city{Nanjing}
  \country{China}
}

\author{Tat-Jen Cham}
\email{astjcham@ntu.edu.sg}
\orcid{0000-0001-5264-2572}
\affiliation{%
  \institution{Nanyang Technological University}
  \city{Singapore}
  \country{Singapore}
}

\author{Jianfei Cai}
\email{jianfei.cai@monash.edu}
\orcid{0000-0002-9444-3763}
\affiliation{%
  \institution{Monash University}
  \city{Melbourne}
  \country{Australia}
}

\begin{abstract}
Although much progress has been made in visual emotion recognition, researchers have realized that modern deep networks tend to exploit dataset characteristics to learn spurious statistical associations between the input and the target. Such dataset characteristics are usually treated as dataset bias, which damages the robustness and generalization performance of these recognition systems. In this work, we scrutinize this problem from the perspective of causal inference, where such dataset characteristic is termed as a \textbf{confounder} which misleads the system to learn the spurious correlation. To alleviate the negative effects brought by the dataset bias, we propose a novel Interventional Emotion Recognition Network (IERN) to achieve the backdoor adjustment, which is one fundamental deconfounding technique in causal inference. Specifically, IERN starts by disentangling the dataset-related context feature from the actual emotion feature, where the former forms the confounder. The emotion feature will then be forced to see each confounder stratum equally before being fed into the classifier.
A series of designed tests validate the efficacy of IERN, and experiments on three emotion benchmarks demonstrate that IERN outperforms state-of-the-art approaches for unbiased visual emotion recognition. Code is available at \url{https://github.com/donydchen/causal_emotion}.
\end{abstract}



\maketitle

%
%
%
%

\section{Introduction}
\begin{figure}[t!]
    \centering
    \includegraphics[width=0.46\textwidth]{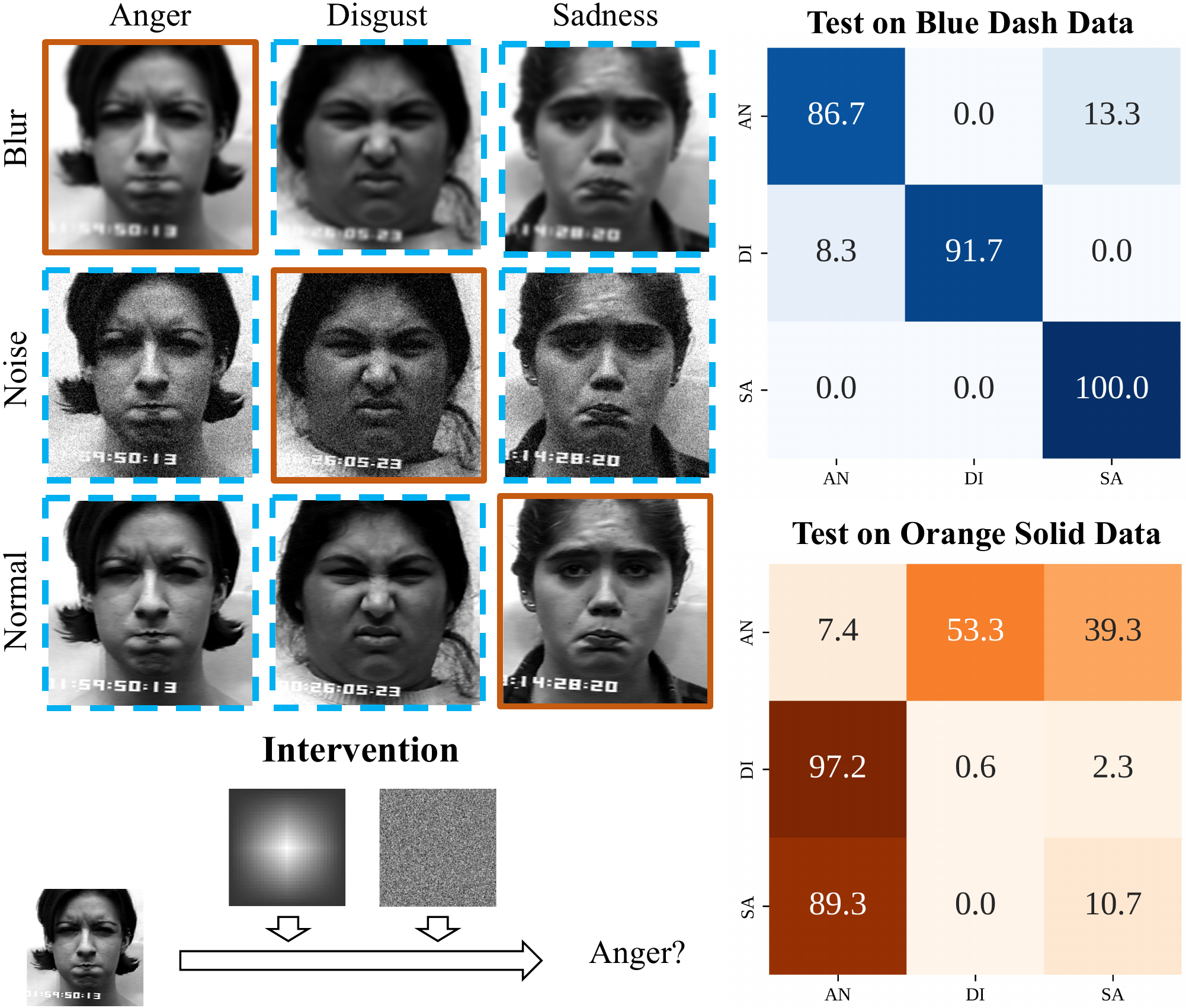}
    \caption{A \emph{toy experiment} to illustrate the dataset bias problem in facial expression recognition, where different expressions are associated with different types of degradation. A model trained on the blue data performed well when tested on other blue samples (blue confusion matrix), but is impaired dramatically when tested on orange samples (orange confusion matrix). The reason is that the model has been confounded to recognize the more easily learned degradation effects. An intuitive solution is via \emph{intervention} by separately adding blur and noise to all input (bottom left).}\label{fig:toy_experiment}
\end{figure}

Visual emotion recognition, including facial expression recognition (FER)~\cite{yan2019joint, lin2020orthogonalization, lee2019context, chen2020label} and image emotion recognition (IER)~\cite{you2015robust, you2016building, panda2018contemplating}, has attracted great attention for decades, playing a vital role in lots of daily scenes such as mental health care and driver drowsiness detection. While deep neural networks (DNNs)  show promising performances on several existing benchmarks~\cite{corneanu2016survey, li2020deep}, it has been recently observed that DNNs may ``cheat'' by relying on unintended cognitive level~\cite{zhao2014exploring} dataset characteristics, \eg, scene contexts~\cite{panda2018contemplating} and human attributes~\cite{wang2020towards}, instead of the actual affective level~\cite{zhao2014exploring} causal relationships between the input image and the output emotion label, harming their effectiveness in practice. Intuitively, such dataset characteristics are leveraged by DNNs as the shortcut~\cite{geirhos2020shortcut} to learn spurious statistical associations between variables.

DNNs may not appear to be affected by dataset characteristics if the training and test set data are 
from the same distribution~\cite{scholkopf2021toward}, under which most emotion benchmarks are collected and constructed. However, a robust practical emotion recognition system is expected to have consistent generalization even for 
data from different distributions~\cite{scholkopf2021toward}. Here we design a \emph{toy experiment} to showcase how DNNs behave on data from both the same distribution and different distributions, as depicted in Fig.~\ref{fig:toy_experiment}. From a facial expression dataset, we rendered additional images by separately applying blur and noise degradation. They were next partitioned into two datasets (blue dashed vs orange solid borders) by selecting subsets of expressions and degradation labels. A VGG-11~\cite{simonyan2014very} was then trained (randomly initialized and without data augmentation) on samples drawn from the blue dataset. 
Such a trained model performed well on samples drawn from the same blue dataset (data from the same distribution), shown in the blue confusion matrix of Fig.~\ref{fig:toy_experiment}, but is impaired dramatically 
when tested on samples from the orange dataset (data from a different distribution). 
Instead of
recognizing expressions based on relevant expression features, the model has been biased towards recognizing the degradation effects, which are easier to detect. 

Unlike existing methods in emotion recognition that address the dataset bias issue via solutions such as re-sampling and building larger datasets, we choose to leverage \textbf{causal inference}~\cite{pearl2000models, pearl2018book, scholkopf2021toward} to tackle it. In causal inference, entities or attributes that mislead DNNs to learn spurious correlations between input and output are termed as \textbf{confounders}~\cite{pearl2000models, pearl2018book, peters2017elements}, which refers to the dataset characteristics in our emotion recognition context. For example, in the above toy experiment, the model has been misled to confound \emph{blur} with \emph{disgust} and \emph{sadness}, and fails to recognize \emph{anger} when \emph{blur} co-occurs. An intuitive solution for the toy experiment is to add noise or blur to all input images (see the bottom left of Fig.~\ref{fig:toy_experiment}), whose underlying principle is to deconfound the emotion recognition systems from the confounder / dataset characteristics. This solution is also known as \textbf{causal intervention}~\cite{pearl2000models}. 

To deal with more sophisticated confounders, \ie, \emph{unknown dataset characteristics} and \emph{image scenes}, we analyze the problem with Pearl's structure causal model (SCM)~\cite{pearl2000models}, and propose a novel framework we call \textbf{Interventional Emotion Recognition Network (IERN)}, to do the intervention by embedding the \textbf{backdoor adjustment} theorem~\cite{pearl2000models}. Although backdoor adjustment has been incorporated in some recent deep learning models for other vision tasks~\cite{yang2020deconfounded, dong2020conta, tang2020longtailed}, all of these only \emph{approximate} the intervention via memory-query operations, rather than applying \emph{real} intervention as is done in our IERN. 

Specifically, our IERN starts by disentangling the dataset-related context feature from the actual emotion feature, where the former represents the confounder.
Following application of the backdoor adjustment theorem, the emotion feature will then be forced to see each confounder stratum equally before being fed into the classifier. We showcase how IERN realizes the backdoor adjustment theorem effectively on the facial expression recognition task via a mixed-dataset configuration, where emotion-independent dataset characteristics are treated as confounder, with dataset name being the label. On the image emotion recognition task, IERN outperforms the state-of-the-art approach by a significant margin under the challenging cross-datset setting, where image scenes are treated as confounders. The contributions are threefold, 
\begin{itemize}
  \item We are the first to tackle dataset bias in visual emotion recognition from a causality perspective.
  \item We propose a novel trainable framework, named Interventional Emotion Recognition Network (IERN), to realize the backdoor adjustment theorem. 
  \item With rigorous experiments done for both the mixed-dataset and cross-dataset on existing benchmarks, we show how IERN effectively alleviates negative effects raised by dataset bias and outperforms state-of-the-art approaches.
\end{itemize}

\section{Related Work}\label{related_work}

\paragraph{Visual Emotion Recognition.} 
The majority of emotion recognition works belong to \textbf{Facial Expression Recognition (FER)}, where the input are human facial images with simple and limited backgrounds~\cite{zhao2011facial,wang2020mead, ji2022eamm}. Most methods focus on enhancing performance by making use of related auxiliary priors, \eg, facial landmarks~\cite{jung2015joint, zhang2017facial, zhang2016deep}, local regions~\cite{jia2019facial, chen2019facial, xie2018facial}, action units~\cite{liu2013aware, chen2022geoconv, chen2020label}, optical flows~\cite{sun2019deep}, \etc. The assumption of these methods is that the training data is fairly distributed, which may be wrong and could harm the generalization performance of trained models. Recently, the bias problem has received much attention, and most approaches~\cite{li2021jdman, mo2021d3net, ruan2021feature, farzaneh2021facial} mainly aim to intuitively disentangle the bias features, so as to build more robust emotion features. 

Others belong to \textbf{Image Emotion Recognition (IER)}, which aims to recognize the emotions invoked from images that contain a variety of objects rather than recognizing emotions of human faces. Approaches along this line mainly deal with social media photos~\cite{wang2015modeling, wu2017inferring}, manually generated images~\cite{jou2014predicting}, artistic photographs~\cite{zhao2014exploring, alameda2016recognizing}, web crawled natural images~\cite{you2015robust, you2016building, panda2018contemplating}, \etc. The most relevant work is~\cite{panda2018contemplating}, 
which identified the image scene as the bias factor, and managed to alleviate the bias issue by building a new larger and more balanced dataset using web data.

Different from all approaches that handle dataset bias in visual emotion recognition via heuristic solutions, our approach is the first to offer a fundamental understanding of the bias issue via causal inference, and propose a new trainable unbiased framework based on causal intervention, by employing the feature disentanglement technique and the center loss function~\cite{wen2016discriminative}.

\paragraph{Causal inference} Causality~\cite{pearl2000models, peters2017elements, scholkopf2021toward} can help pursue the causal effect between two observed variables, rather than depend only on their correlation. Many works have been proposed to explore how causality can be leveraged in machine learning~\cite{magliacane2018domain, parascandolo2018learning, besserve2018counterfactuals, bengio2019meta}. Recently, causal inference has been introduced into the deep learning framework for several computer vision tasks, including image captioning~\cite{yang2020deconfounded}, image classification~\cite{chalupka2015visual, lopez2017discovering, tang2020longtailed}, semantic segmentation~\cite{dong2020conta}, and few-shot learning~\cite{yue2020interventional}. 

Our work differs from  causality-based deep learning methods in two aspects.
Firstly, ours is the first to address the biased visual emotion recognition problem. Secondly and more importantly, we propose a novel solution by learning confounder features and conducting \emph{real} intervention, while previous methods had chosen to model confounders using a predefined dictionary and approximate intervention via memory-query operations by adopting the Normalized Weighted Geometric Mean (NWGM)~\cite{baldi2014dropout, xu2015show}.

\section{Methodology}\label{method}

Given input image $X$, visual emotion recognition aims to solve the problem $P(Y | X)$, where $Y$ is the emotion label. 
In this section, we will detail how confounder $D$ undermines the objective of $P(Y | X)$ (Section~\ref{method:scm}) to raise dataset bias, how such a bias effect can be removed (Section~\ref{method:backdoor}), and what our unbiased solution is (Section~\ref{method:implementation}).

\subsection{SCM for Emotion Recognition}\label{method:scm}
\begin{figure}[t!]
    \centering
    \includegraphics[width=0.47\textwidth]{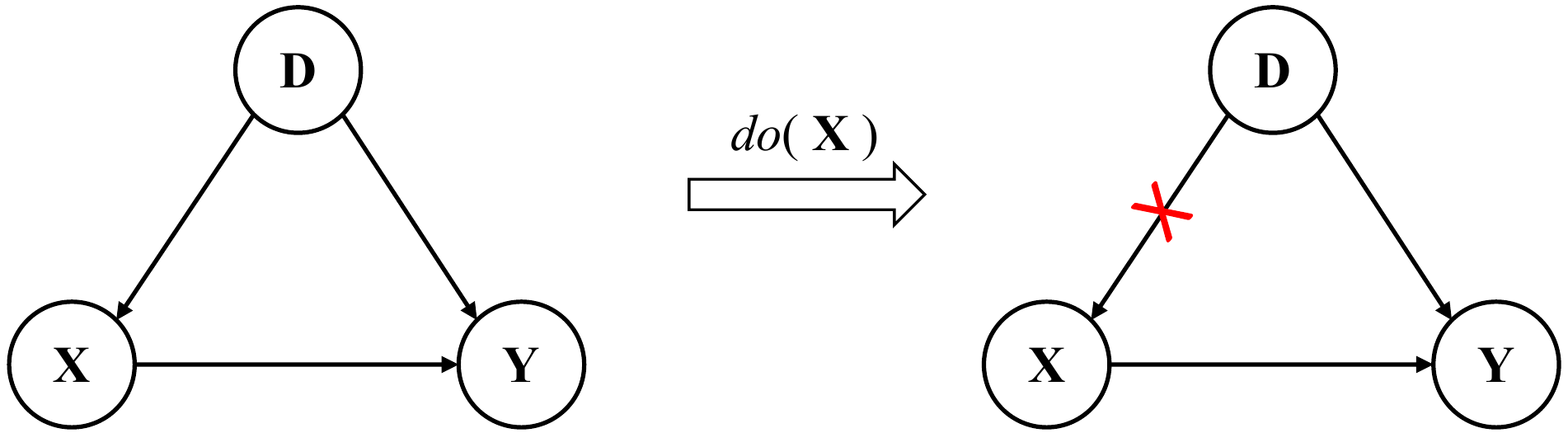}
    \caption{Left: structural casual model (SCM) for visual emotion recognition. Right: intervention with backdoor adjustment. The direction  of  an  edge  denotes  only  the causal relationship, pointing from the cause to the effect, while information can flow bidirectionally. X, Y, D refer to input image, emotion label and confounder, respectively.}\label{fig:scm}
\end{figure}

Structural causal models (SCM)~\cite{pearl2000models, scholkopf2021toward} are designed to analyze the causal relationships. 
Fig.~\ref{fig:scm} shows a SCM in our context constructed among input $X$, emotion $Y$ and confounder $D$. The direction of an edge denotes only the causal relationship, while information can flow bidirectionally, \eg, $D \rightarrow X$ means that $D$ is the cause and $X$ is the effect, while information can still flow from $X$ to $D$.

\paragraph{$\boldsymbol{X \rightarrow Y}$.} It is the intended causal relationship that the network is supposed to learn, which is to recognize emotion $Y$ based on input image $X$. 
For simplicity, we mix the use of symbol $X$, denoting either input images or image features.

\paragraph{$\boldsymbol{D \rightarrow X,}$ $\boldsymbol{D \rightarrow Y}$.} Confounder $D$ is the undesirable context feature, \eg, \textit{unknown dataset characteristics, image scenes}. Without disentanglement, confounder are normally embedded in the input features, and thus we have $D \rightarrow X$. 
Similarly, confounder also affects networks in predicting emotion, so we have $D \rightarrow Y$.

\paragraph{$\boldsymbol{X \leftarrow D \rightarrow Y}$ \textbf{(backdoor path)}.} 
The existence of confounder $D$ enables a backdoor path between $X$ and $Y$, by which networks can be biased to build up spurious correlation between $X$ and $Y$, leading to degradation of its generalization performance. For example, as shown in Fig.~\ref{fig:toy_experiment}, instead of identifying the correct \emph{anger} image features to recognize \emph{anger}, \ie $X \rightarrow Y$, networks may learn that \emph{noise} is \emph{anger}, \ie $D \rightarrow X$, and should be predicted as \emph{anger}, \ie $X \leftarrow D \rightarrow Y$.

The backdoor path can be explained with the law of total probability. Specifically, $P(Y | X)$ can be decomposed into
\begin{equation}
P(Y|X) = \sum_{d}^{}P(Y|X, D=d)P(D=d|X), 
\end{equation}
where $d$ is one of the strata (a.k.a. ``level'') of confounder $D$, \eg, $d_{  \mathrm{blur} }$, $d_{ \mathrm{noise}}$. Assuming that spurious correlation leads to $P(d_{\mathrm{noise}} | X_{\mathrm{anger}}) \approx 1$, then $P(Y|X_{\mathrm{anger}})$ will become $P(Y|X_{\mathrm{anger}}, d_{\mathrm{noise}})$, resulting in bias towards $d_{\mathrm{noise}}$.

\begin{figure*}[t!]
    \centering
    \includegraphics[width=0.97\textwidth]{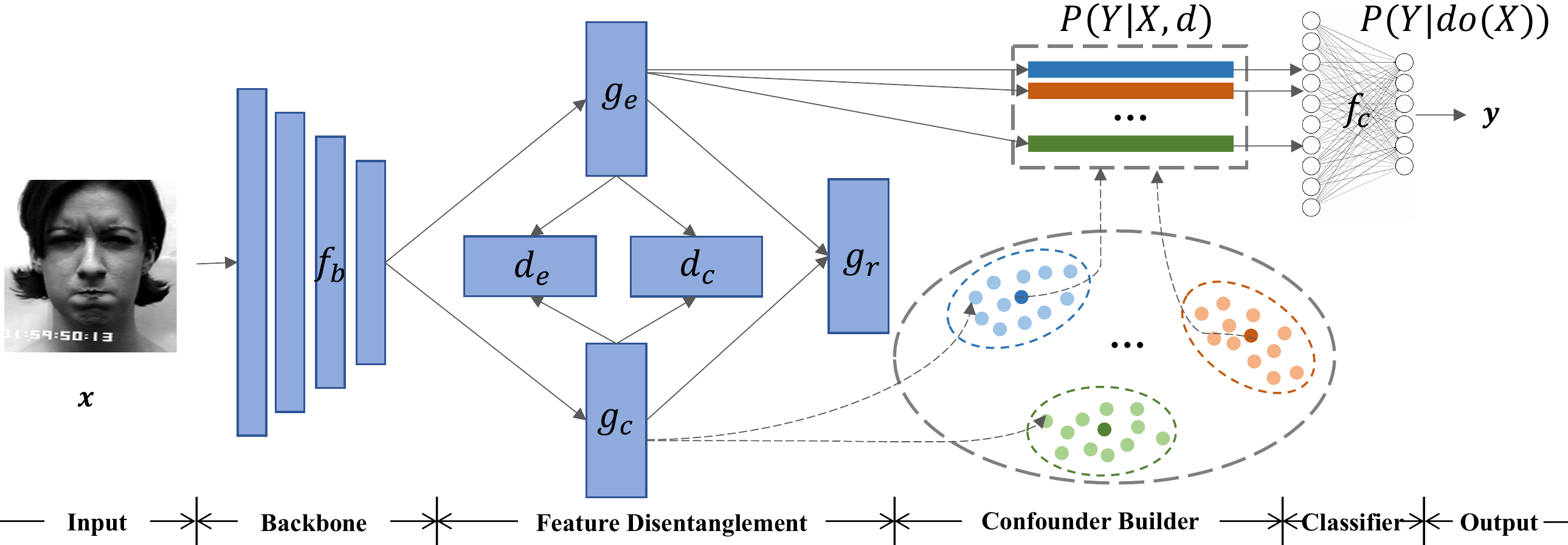}
    \caption{Overview of Interventional Emotion Recognition Network (IERN). $f_b$, $g_e$, $d_e$, $g_c$, $d_c$, $g_r$ and $f_c$ refer to backbone, emotion generator, emotion discriminator, context generator, context discriminator, reconstruction network and classifier, respectively. In training, Feature Disentanglement splits features into emotion and context features. Confounder Builder groups context features and gets their centers as confounder features. Each emotion feature is combined with all strata of the confounder individually, and forwarded to the Classifier. Only the top branch will be leveraged in testing, using the learned confounders.} \label{fig:dnn_model}
\end{figure*}

\subsection{Intervention with Backdoor Adjustment}\label{method:backdoor}
Once the confounder is identified, a straightforward solution to alleviate the dataset bias is to build a larger dataset to ensure that emotion is balanced among all confounder strata, similar to what has been done in \cite{panda2018contemplating}. However, such solutions might not be practical due to three main reasons. Firstly, it is time-consuming and costly to collect and label a large amount of data. Secondly, if the confounder is too complex, it is hard to construct a balanced new dataset. Thirdly, there may be ethical or practical issues in collecting certain types of data, \eg, forcing a child to cry in a park in order to get negative-emotional park images.

We therefore turn to a more elegant solution of applying the \textbf{intervention}~\cite{pearl2000models} by blocking the backdoor path in the SCM (Fig.~\ref{fig:scm} Right). Specifically, instead of learning the correlation $P(Y|X)$, we set our target as $P(Y| do(X))$, where $do(\cdot)$ is the \textbf{\textit{do}-operator} (a.k.a. \textit{do}-calculus), denoting the experimental intervention. 
Since both $D$ and $X$ are accessible in our tasks, we can solve $P(Y| do(X))$ by applying the backdoor adjustment theorem~\cite{pearl2000models}:
\begin{equation}
    P(Y| do(X)) = \sum_{d}^{}P(Y|X, D=d)P(D=d).
\end{equation}
Intuitively, the backdoor adjustment removes the risk of spurious correlation arising from $P(D|X)$ via a two-step strategy. It starts by estimating the causal effect in each stratum of the confounder, followed by summing the effects of all strata, weighted by their known prevalence in the population. By doing so, it ensures that the contribution of the causal effect of an \emph{unobserved} stratum is no different from that of an \emph{observed} one. In this way, the overall causal effect would not be biased towards the observed strata. For example, with the help of the backdoor adjustment, each stratum of the confounder, including \textit{noise}, \textit{blur} and \textit{normal}, will contribute equally to \textit{anger}, forcing the model to learn to recognize \textit{anger} from the intended affective features.

\subsection{IERN}\label{method:implementation}

To overcome dataset bias in visual emotion recognition, we propose our Interventional Emotion Recognition Network (IERN) by embedding the backdoor adjustment. Compared with heuristic debiasing methods like building larger datasets~\cite{panda2018contemplating}, IERN is designed based on well-studied deconfounding theory. Also, unlike NWGM approximation-based backdoor adjustment models~\cite{dong2020conta, yang2020deconfounded} that absorb the probability sum into the network, IERN applies the real intervention $P(Y|do(X))$.
Fig.~\ref{fig:dnn_model} gives an overview of the IERN, which is composed of four parts: Backbone, Feature Disentanglement, Confounder Builder and Classifier. The backbone refers to an existing image feature extractor, \eg, VGG~\cite{simonyan2014very} or ResNet~\cite{he2016deep}, the disentanglement part separates emotion and context features, the confounder builder does the causal intervention, and the classifier predicts the emotion label. Formally, $f_b$, $g_e$, $d_e$, $g_c$, $d_c$, $g_r$ and $f_c$ refer to backbone, emotion generator, emotion discriminator, context generator, context discriminator, reconstruction network and classifier, respectively.  

\paragraph{Feature Disentanglement.} As mentioned in Section~\ref{method:scm}, without proper handling, context features are mixed with emotion features. Thus, IERN starts by applying feature disentanglement to separate emotion and context features. Specifically, to obtain the clean emotion feature, the disentanglement is conducted by a dual-discriminator structure, consisting of an emotion discriminator $d_e$ and a context discriminator $d_c$, where $d_e$ is to ensure the extracted feature contains emotional information while $d_c$ is to ensure that the extracted feature does not contain any context information.

This procedure is implemented by alternating the optimization between the feature generators and the discriminators with the following training objective:
\begin{equation}\label{eq:le}
\begin{aligned}
    \mathcal{L}_{e} &= \min_{d_e^{\theta}} l_{\textrm{CE}}(d_e(g_e(f_b(x))), y_e) \\
    &\quad + \min_{g_e^{\theta}} l_{\textrm{MSE}}(d_c(g_e(f_b(x))), \frac{1}{N_c}), 
\end{aligned}
\end{equation}
where $l_{\textrm{CE}}$ and $l_{\textrm{MSE}}$ denote the cross-entropy loss and the mean squared error loss, respectively; $d_e^{\theta}$ and $g_e^{\theta}$ are the trainable parameters of $d_e$ and $g_e$, respectively; and $g_e(f_b(x))$ is the intended clean emotion feature, obtained by feeding the input image $x$ first into the backbone $f_b$ and then into the emotion feature generator $g_e$. Also, $y_e$ is the ground-truth emotion label for $x$, and $N_c$ is the number of levels in the confounder. Intuitively, in Eq.~\eqref{eq:le}, $d_e$ is optimized to predict the ground truth emotion label given the generated emotion feature $g_e(f_b(x))$ as input, so as to ensure that $g_e(f_b(x))$ contains the correct emotion information.
Here, $g_e$ is optimized to encourage the softmax output of $d_c$ be equally distributed among all confounder levels, so as to ensure that $g_e(f_b(x))$ contains no context information. 

To obtain clean context features, we adopt a similar optimization approach with the following objective function:
\begin{equation}
\begin{aligned}
    \mathcal{L}_{c} &= \min_{d_c^{\theta}} l_{\textrm{CE}}(d_c(g_c(f_b(x))), y_c) \\
    &\quad + \min_{g_c^{\theta}} l_{\textrm{MSE}}(d_e(g_c(f_b(x))), \frac{1}{N_e}),
\end{aligned}
\end{equation}
where $d_c^{\theta}$ and $g_c^{\theta}$ are the trainable parameters of the context discriminator $d_c$ and the context generator $g_c$, respectively; $y_c$ is the confounder label, and $N_e$ is the number of emotion classes. 

To ensure that the separated features fall within reasonable domains, IERN should be capable of reconstructing the base feature $f_b(x)$, given the emotion features and the context features as input. Thus, a feature reconstruction loss function is added as
\begin{equation}
    \mathcal{L}_{r} = \min_{g_r^{\theta}, g_e^{\theta}, g_c^{\theta}} l_{\textrm{MSE}} (g_r(g_e(f_b(x)), g_c(f_b(x))), f_b(x)),
\end{equation}
where $g_r^{\theta}$ is the trainable parameters of the feature reconstruction network $g_r$.

\paragraph{Confounder Builder.} The purpose of the confounder builder is to combine each emotion feature with different context features so as to avoid the bias towards the observed context strata. To limit the diversity of the context features, we propose to use the \textit{center} of all context features within each specific stratum as a confounder feature. This is reasonable since   
context features are usually similar within the same stratum while different across different strata. Essentially, a confounder feature represents the \textit{general concept} of a specific stratum. So we adopt the center loss~\cite{wen2016discriminative} to learn confounder features with the objective function as
\begin{equation}
    \mathcal{L}_{\textrm{CB}} = \min_{\mathcal{C},g_c^{\theta}} l_{\textrm{MSE}} (g_c(f_b(x)), \mathcal{C}_{y_c}),
\end{equation}
where $\mathcal{C}_{y_c}$ denotes the learned confounder feature of the $y_c$-th stratum of the confounder, and $\mathcal{C} = \{ \mathcal{C}_1, \mathcal{C}_2, ..., \mathcal{C}_{N_c}\}$, with all being learnable parameters.

\paragraph{Classifier.} 
As mentioned in Section~\ref{method:backdoor}, the backdoor adjustment aims to weigh the causal effect of each stratum with its prevalence in the population, \ie, $P(D=d)$, instead of $P(D=d|X)$.
To ensure fairness, our deconfounded classifier is set such that for any emotion feature, it is present in all confounder strata equally, \ie, $P(D=d) = 1/N_c$, via the following training objective:
\begin{equation}
     \mathcal{L}_{\textrm{Cls}} = \min_{f_c^{\theta}, g_e^{\theta}, f_b^{\theta}} l_{\textrm{CE}} ( \frac{1}{N_c} \sum_{i=1}^{N_c} f_c(g_r( g_e(f_b(x)), \mathcal{C}_i) ), y_e), 
\end{equation}
where $f_c^{\theta}$ denotes the trainable parameters of the classification network $f_c$. Given a separated emotion feature $g_e(f_b(x))$, IERN reuses the reconstruction network $g_r$ to combine it with each learned confounder feature $\mathcal{C}_i$, so as to ensure that for any emotion, the classifier can be challenged with that emotion existing in every stratum of the confounder equally, thus avoiding bias. In other words, $P(Y|X, D=d)$ is obtained by $f_c(g_r( g_e(f_b(x)), \mathcal{C}_i) )$, while by forwarding $N_c$ times and taking the average, we have $P(Y|do(X))$, where $P(D=d) = 1/N_c$.

\paragraph{Training Phase.} 
In general, IERN is trained with the weighted combination of the above loss functions. The overall objective 
is
\begin{equation}\label{eq:ldo}
\mathcal{L}_{\textit{do}} = \lambda_1 (\mathcal{L}_{e} + \mathcal{L}_{c} + \mathcal{L}_{r}) +  \lambda_2  \mathcal{L}_{\textrm{CB}} + \lambda_3  \mathcal{L}_{\textrm{Cls}}, 
\end{equation}
where $\lambda_1$, $\lambda_2$ and $\lambda_3$ are hyper-parameters. IERN is end-to-end trainable, and different components need to be updated sequentially within one training iteration. The detailed training procedure is given in Algorithm~\ref{alg:train} (see Section~\ref{sec:app_alg} for technical details).
\begin{algorithm}
\caption{IERN Training Procedure} \label{alg:train}
\begin{algorithmic}[1]
\For{number of training iterations}
    \State Forward $f_b$, $g_e$, $d_e$, $g_c$, $d_c$, $g_r$
    
    \State Freeze $g_e^\theta$ and $g_c^\theta$, backward $d_e$ using 1st term of $\mathcal{L}_e$ and $d_c$ using 1st term of $\mathcal{L}_c$
    
    \State Freeze $d_e^\theta$ and $d_c^\theta$, backward $g_e$ using 2nd term of $\mathcal{L}_e$ and $\mathcal{L}_r$; backward $g_c$ using 2nd term of $\mathcal{L}_c$, $\mathcal{L}_r$ and $\mathcal{L}_{\textrm{CB}}$; backward $g_r$ using $\mathcal{L}_r$; backward $\mathcal{C}$ using $\mathcal{L}_{\textrm{CB}}$

    \For{$i=1\rightarrow N_c$}
        \State Forward $f_c(g_r( g_e(f_b(x)), \mathcal{C}_i))$
    \EndFor
    \State Backward $f_c$, $g_e$ and $f_b$ using $\mathcal{L}_{\textrm{Cls}}$
\EndFor
\end{algorithmic}
\end{algorithm}

\paragraph{Testing Phase.} 
Given a test image, IERN will predict its emotion by going through only the top branch (see Fig.~\ref{fig:dnn_model}), which is 
\begin{equation}
    \hat{y_e} = \mathop{\arg\max} \sigma( \frac{1}{N_c} \sum_{i=1}^{N_c} f_c(g_r( g_e(f_b(x)), \mathcal{C}_i) ) ),
\end{equation}
where $\sigma$ denotes the softmax function, and $\mathcal{C}_i$ is the confounder feature learned in the training phase. 
Following the backdoor adjustment theorem, to ensure the prediction of emotion is not biased towards specific strata at test time, we need to combine the emotion feature with each confounder stratum feature and take an average.

\section{Experiments}\label{experiment}

We conducted experiments on both facial expression recognition (FER) (Section~\ref{exp:fer}) and image emotion recognition (IER) (Section~\ref{exp:ier}). For the former, we designed a challenging yet practical out-of-distribution (o.o.d.\ ) test configuration through mixing different benchmarks, \emph{motivated by the fact that in the real world, images provided to practical recognition systems are taken by different people with different preferences, cameras and geographical locations}, and this distribution cannot be controlled~\cite{scholkopf2021toward}. 
And we conduct ablation study and compare to related SOTA, including DACL~\cite{farzaneh2021facial} and NWGM-based methods~\cite{dong2020conta}, using the introduced configurations.
As for the latter, IERN is compared, under the challenging cross-dataset setting, to Curriculum Learning~\cite{panda2018contemplating}, which is the SOTA for unbiased image emotion recognition.

\paragraph{Implementation details.}
To maintain a fair comparison, ResNet-50~\cite{he2016deep} was chosen as the backbone $f_b$ following the setting of~\cite{panda2018contemplating}. While $g_e$, $g_c$ and $g_r$ are constructed by using the residual block~\cite{he2016deep} as building blocks, $d_e$ and $d_c$ are composed of convolution layers, and $f_c$ is composed of fully connected layers (see Section~\ref{sec:app_network} for more details). 
Note that other backbones or building blocks can also be adopted, as long as they follow the framework depicted in Fig.~\ref{fig:dnn_model}. The hyper-parameters $\lambda_1$ and $\lambda_3$ in Eq.~\eqref{eq:ldo} were set as default to $\lambda_1$=$\lambda_3$=1, while $\lambda_2$=5$\times 10^{-4}$ were set mainly to balance the loss value magnitudes (see Section~\ref{sec:app_hyper} for more details).
We initialized $f_b$ with ImageNet~\cite{deng2009imagenet} pretrained weights, while other components were randomly initialized. An Adam optimizer was used with a learning rate of 2$\times 10^{-4}$, decayed using the standard warm-up strategy.  All models were trained till converged before testing. Specifically, in the mix-dataset experiments (Section~\ref{exp:fer}), IERN were trained for 80 epochs, while in the cross-dataset experiments (Section~\ref{exp:ier}), IERN were trained for 140 epochs. All test sets remain untouched during training.
We implemented IERN with PyTorch~\cite{paszke2019pytorch}. 

\begin{table}[t!]
\centering
\caption{Three-fold cross-validation setting. For each fold, image sets denoted with the corresponding number are selected as the test set, with the others as the training set}
\begin{tabular*}{.478\textwidth}{@{\extracolsep{\fill}}|l|c|c|c|c|c|c|}
\hline
           & AN   & DI   & FE   & HA   & SA   & SU   \\ \hline
CK+        & $1$  & $2$  & $3$  & $1$  & $2$  & $3$  \\ \hline
MMI        & $3$  & $1$  & $2$  & $3$  & $1$  & $2$  \\ \hline
Oulu-CASIA & $2$  & $3$  & $1$  & $2$  & $3$  & $1$  \\ \hline
\end{tabular*}
\label{tab:three_folds}
\end{table}

\subsection{Mix-Dataset Experiments for FER}\label{exp:fer}

\paragraph{Dataset settings.}
The facial expression experiments were conducted on three lab-controlled benchmarks, including CK+~\cite{lucey2010extended}, MMI~\cite{valstar2010induced}, and Oulu-CASIA~\cite{zhao2011facial}. In all the datasets, we only selected those sequences that contain frontal faces and are labeled with six basic expression labels, \ie, \textit{anger}, \textit{disgust}, \textit{fear}, \textit{happiness}, \textit{sadness} and \textit{surprise}, resulting in CK+ with 327 sequences of 118 subjects, MMI with 208 sequences of 32 subjects, and Oulu-CASIA with 480 sequences of 80 objects. As a general approach~\cite{li2020deep}, three peak frames were extracted from each sequence for further experiments.

\begin{table*}[t!]
\centering
\caption{FER accuracy for the comparisons using the \emph{designed o.o.d. setting}. The best results are highlighted in bold} 
\begin{tabular}{@{}lccccccc@{}}
\toprule
 & Anger & Disgust & Fear & Happiness & Sadness & Surprise & Average \\ \midrule
Baseline    &\textbf{10.13}          & 30.60 & 15.29          & 49.56          & 10.00          & 77.12          & 35.57          \\
Disentanglement & 1.27           & 27.68          & 18.80          & 67.19          & 15.95          & 80.07          & 38.95          \\
DACL~\cite{farzaneh2021facial} & 5.27 & \textbf{30.68} & \textbf{29.82} & 47.99 & 11.90 & \textbf{83.50} & 38.01 \\
IERN (Ours) & \textbf{10.13} & 28.07          & 23.56 & \textbf{81.85} & \textbf{29.05} & 79.09 & \textbf{45.50} \\  \bottomrule
\end{tabular}
\label{tab:fer_results}
\end{table*}

To ascertain the effectiveness of IERN in dealing with dataset bias, practical o.o.d.\ tests were designed by mixing the three datasets in a three-fold cross-validation setting. With reference to Tab.~\ref{tab:three_folds}: in the first fold, image sets denoted `1', \eg, all \emph{anger} images from CK+, were included in the test set, while the other images were used to construct the training set. Likewise, the second fold comprises a test set of the `2' image sets, and so on.
Here we chose to apply a sequence-independent setting rather than a person-independent one, the reason is that using the latter will make the data very unbalanced across different folds, 
since the numbers of emotions vary across different people on CK+ and MMI.
The performance was measured in terms of classification accuracy, averaged across all three folds (\% is omitted in all result tables). We point out that under such a setting, the dataset label is the confounder label, \ie, $y_c \in \{ \textrm{CK+}, \textrm{MMI}, \textrm{Oulu-CASIA} \}$. Note that we did not specify
which parts of the dataset features were biased, but treated all features that can identify each specific dataset jointly as the confounder. This is reasonable since dataset bias in practice can be very complex and it is often hard to identify different bias factors.

\begin{figure}[t!]
    \centering
    \includegraphics[width=0.265\textwidth]{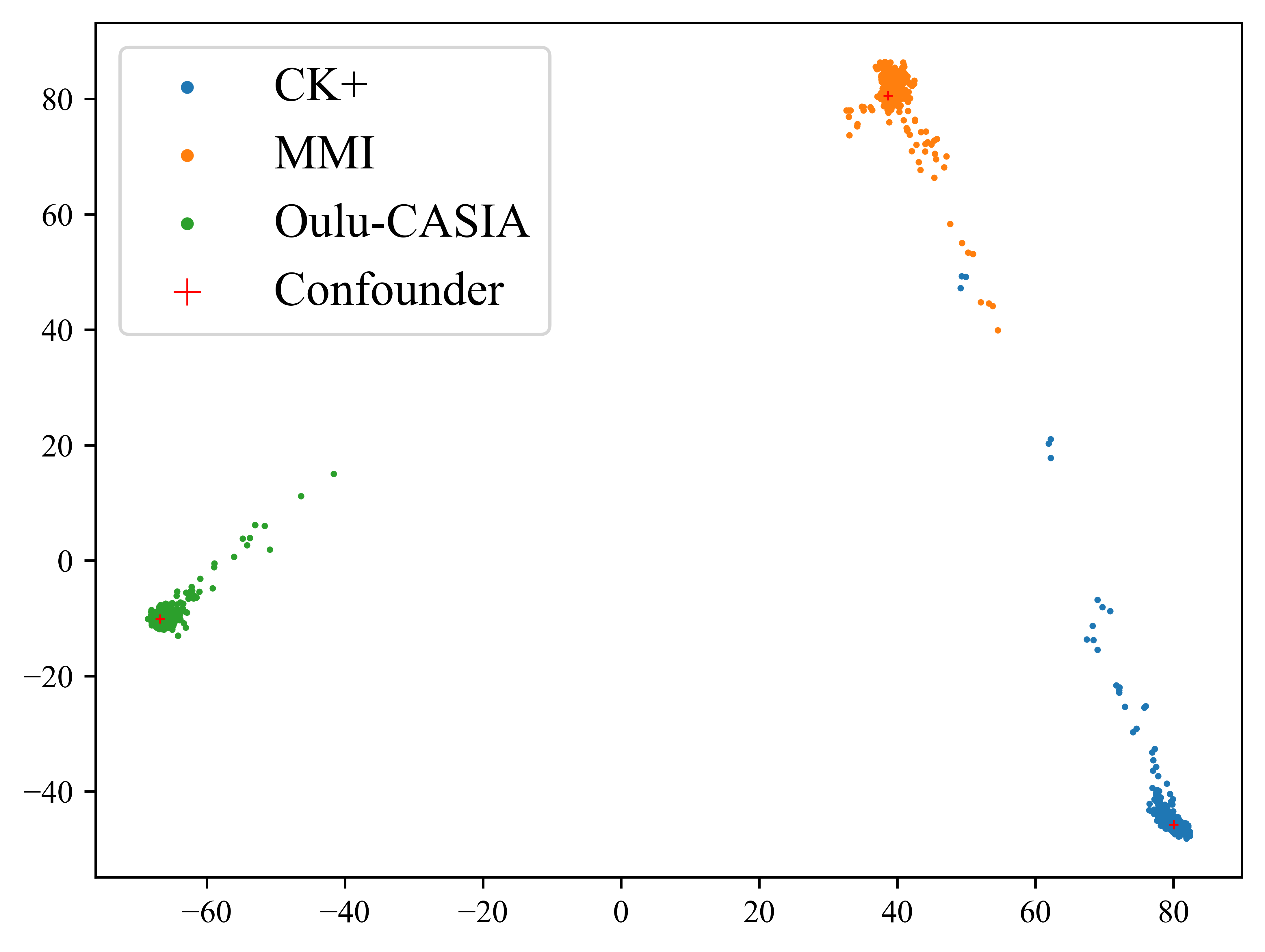} 
    \includegraphics[width=0.195\textwidth]{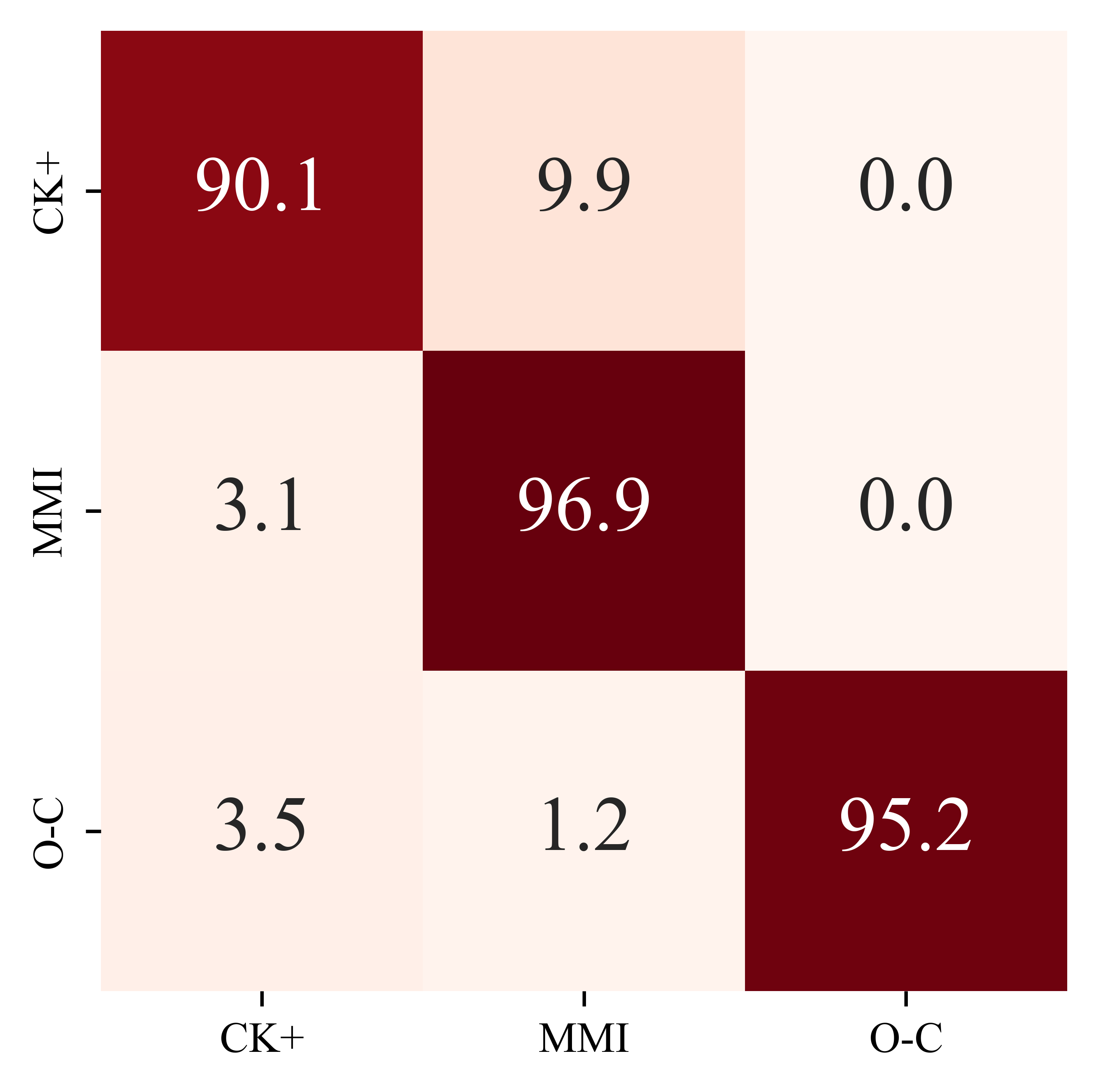}
    \caption{Left: t-SNE maps of context and confounder features (best viewed in high-resolution color). Right: confusion matrix of the confounder prediction.}
    \label{fig:fer_cnfnd}
\end{figure}

\paragraph{Ablation study of IERN}
We compared IERN to two variants: 1) Baseline, which is the backbone of IERN, \ie, ResNet-50. 2) Disentanglement, which directly feeds the output of $g_e$ into $f_c$, bypassing the Confounder Builder. The results are shown in Tab.~\ref{tab:fer_results}.

Compared to Baseline, Disentanglement saw a 3.38\% improvement in terms of average accuracy. This is because Disentanglement can distill the expected emotion feature whilst dispelling the undesirable confounding feature, so that the classifier can be better learned. This finding has also been explored and verified recently by an existing approach that dealt with biased facial attribute classiﬁcation \cite{alvi2018turning}. IERN improved on Disentanglement by 6.55\%. 
From a causality point of view, IERN invokes causal intervention, so that it can address the bias better as previously discussed. From a deep learning point of view, IERN can be seen as applying \emph{feature-level data augmentation}. It combines emotion features with all kinds of confounder features to provide coverage even for combinations missing from training data. So it encounters greater data diversity and is thus more robust than pure disentanglement. 

We also evaluated the related SOTA, DACL~\cite{farzaneh2021facial}, on the designed setting by training and testing with the released codes. As shown in Tab.~\ref{tab:fer_results}, IERN outperformed DACL with a healthy margin. Besides, the performance of DACL is close to Disentanglement, since they share similar underlying principles as mentioned in Section~\ref{related_work}.

\begin{figure}[t!]
    \centering
    \includegraphics[width=0.22\textwidth]{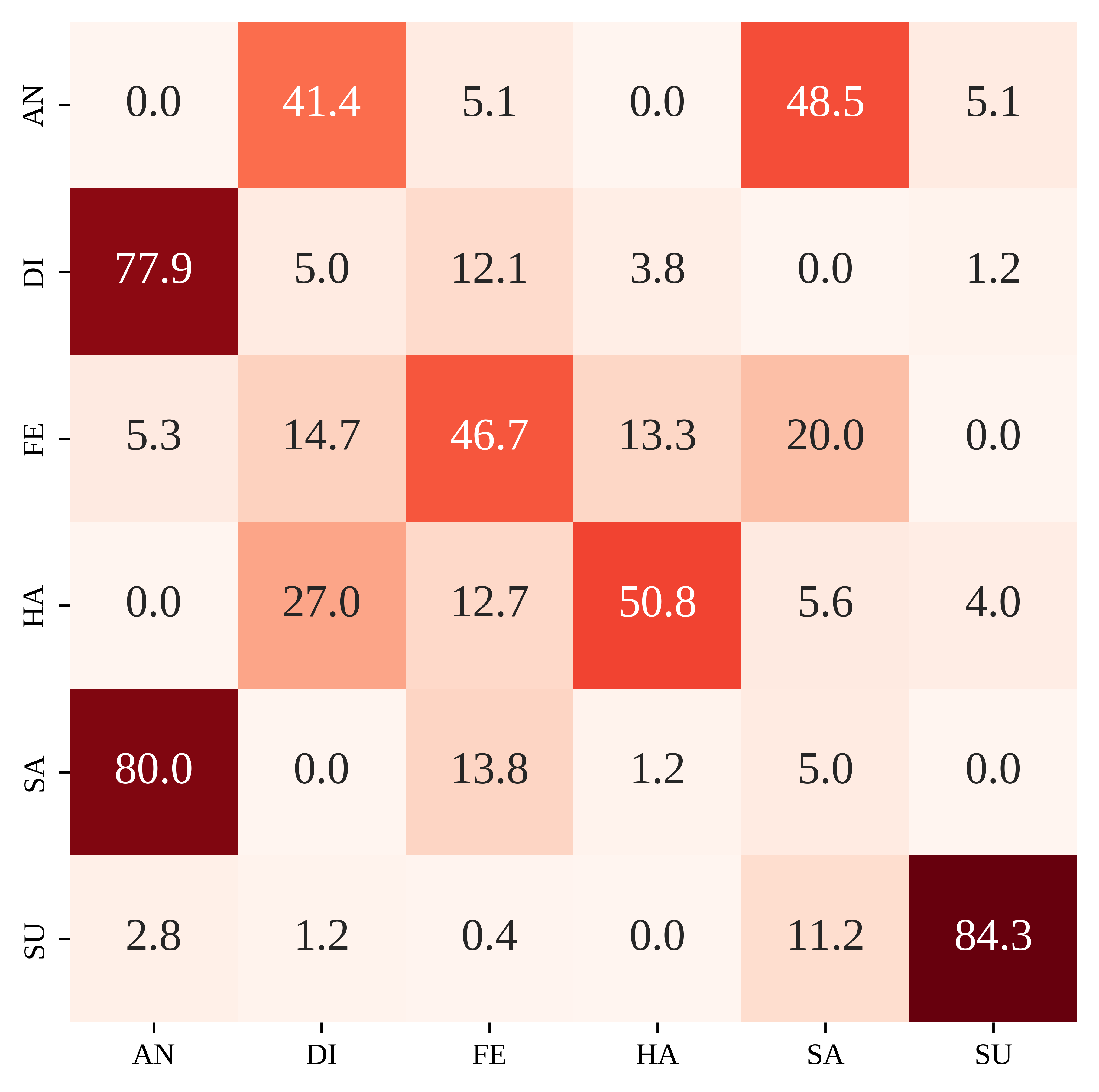} 
    \quad
    \includegraphics[width=0.22\textwidth]{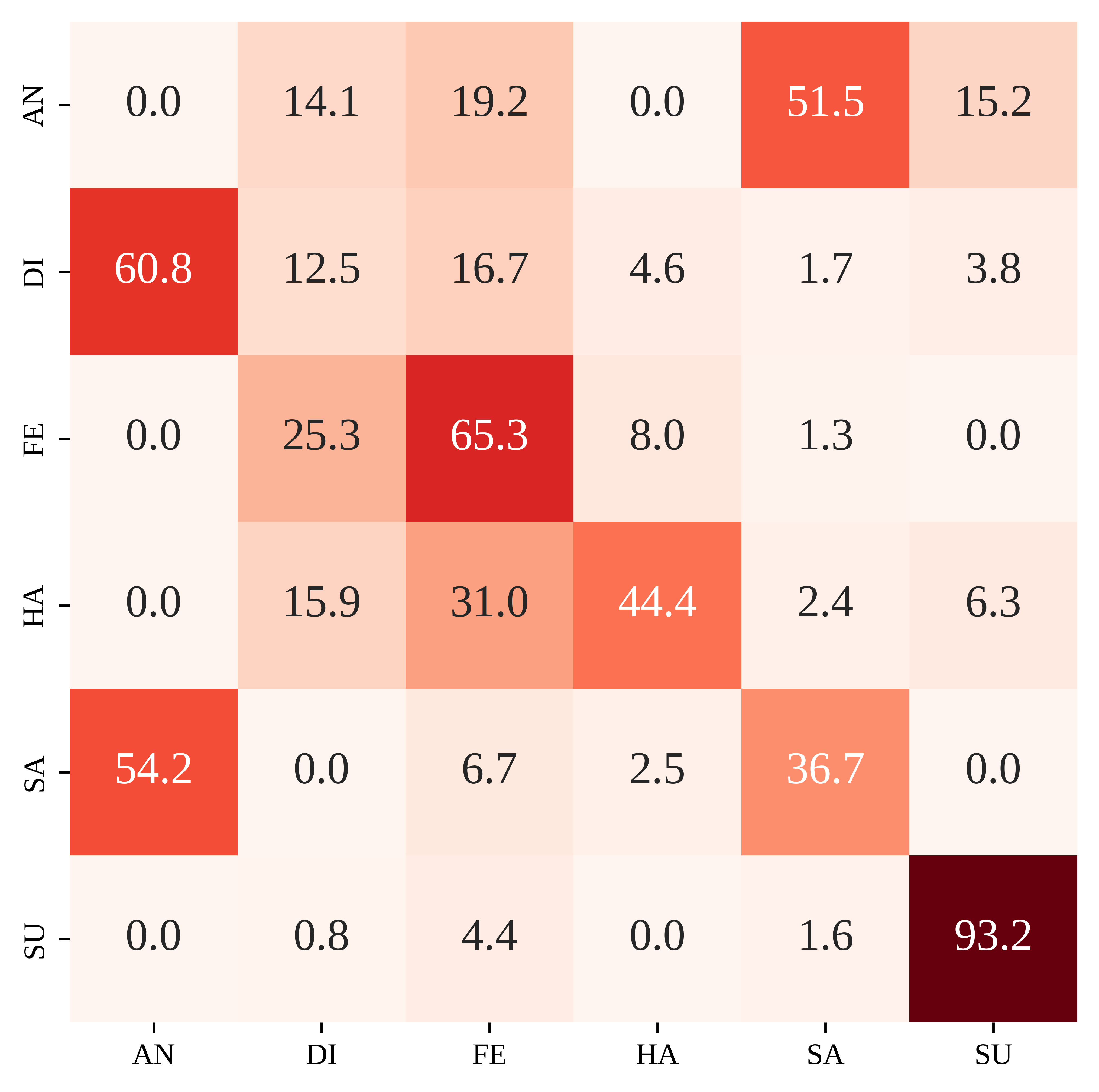} 
    \caption{Confusion matrices of accuracy for both Baseline (left) and IERN (right) on the third fold setting.}
    \label{fig:fer_ecm}
\end{figure}

\paragraph{Confounder features are correctly modeled.}
We plotted the extracted context features $g_c(f_b(x))$ and confounder center features $\mathcal{C}$ using t-SNE. As seen in the left part of Fig.~\ref{fig:fer_cnfnd}, context features were grouped and separated, while confounder features were found near the center of each context feature distribution as expected. Therefore, by using the Confounder Builder, the feature of each stratum of the confounder is correctly modeled, allowing the following \textit{do} intervention to be applied correctly.

\paragraph{Dataset is the confounder.}
The right part of Fig.~\ref{fig:fer_cnfnd} depicts the accuracy confusion matrix for $d_c$, given the extracted context feature $g_c(f_b(x))$ as input.
We can see that the network achieved over 90\% accuracy in predicting the confounder / dataset label, which is consistent with the finding of the \textit{Name That Dataset Game} conducted in both \cite{torralba2011unbiased} and \cite{panda2018contemplating}. Compared with the emotion prediction results shown in Tab.~\ref{tab:fer_results}, it is clear that the network is better at learning to recognize the cognitive level features (dataset characteristics) than to recognize the affective level features (image emotion). Without intervention, the network would thus be biased by the more easily learned dataset features.

\begin{table*}[t!]
\centering
\caption{Facial expression recognition accuracy for the comparison with other methods using the \emph{modified setting} (moving 10\% of images from the original test set into the training set). The best results for each category are highlighted in bold} 
\begin{tabular}{@{}lccccccc@{}}
\toprule
& Anger & Disgust & Fear & Happiness & Sadness & Surprise & Average \\ \midrule
Baseline    & 24.48          & 46.10          & 49.44          & 80.04          & 46.83          & 89.31          & 58.55          \\
Re-sampling  & 23.78          & 48.49          & 57.78          & \textbf{89.73} & 43.92          & 90.22          & 61.59          \\
NWGM-based~\cite{dong2020conta}  & 30.77          & \textbf{59.74}          & 52.78          & 81.59 & 45.24          & \textbf{95.29}          & 63.63          \\
IERN (Ours) & \textbf{59.44} & 59.52 & \textbf{60.56} & 83.92          & \textbf{62.43} & 93.66 & \textbf{71.71} \\ \bottomrule
\end{tabular}
\label{tab:fer_results_updated}
\end{table*}

\paragraph{IERN vs Baseline in detail.}
To get a closer view of how IERN outperforms Baseline in dealing with dataset bias, the accuracy confusion matrices of both models on the third fold setting are shown in Fig.~\ref{fig:fer_ecm}.
Compared to IERN, the results of Baseline were biased by the training data. For example, \textit{fear} images in the test set are from CK+ (see Tab.~\ref{tab:three_folds} number $3$), Baseline predicted over 50\% images as \textit{anger}, \textit{disgust}, \textit{happiness} or \textit{sadness}, which are all related to CK+ in the training set, while IERN alleviated the bias and improved the performance by near 20\%. 

\begin{figure}[t!]
    \centering
    \includegraphics[width=0.475\textwidth]{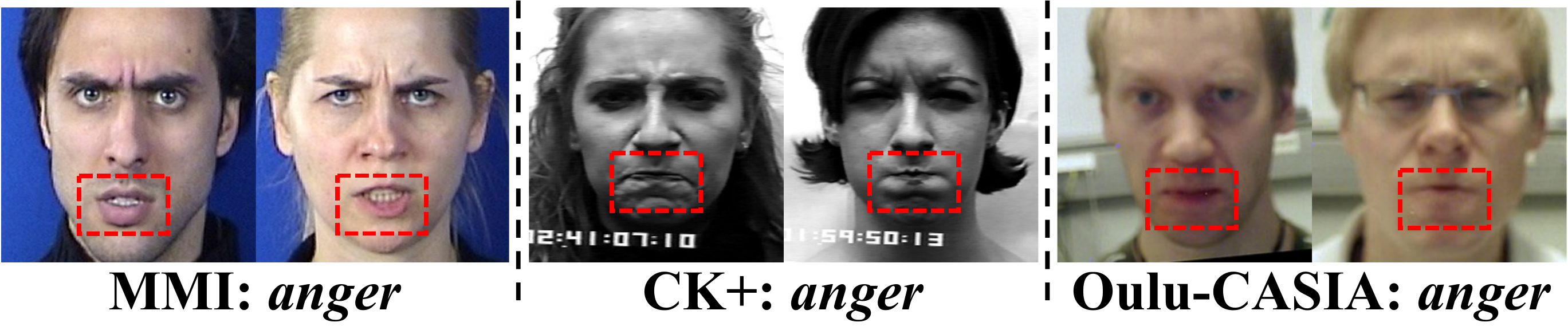}
    \caption{Two samples of \textit{anger} from each dataset. See the mouth region, while it is consistently opened within MMI, such manifestation is inconsistent in other two datasets.}\label{fig:fer_fail}
\end{figure}

\paragraph{Limitation of using dataset as the confounder.}
We noticed that the accuracy results of \textit{anger} for both Baseline and IERN are 0 as shown in Fig.~\ref{fig:fer_ecm}. The reason is that the manifestation of \textit{anger} in MMI is inconsistent with that in CK+ and Oulu-CASIA, as depicted in Fig.~\ref{fig:fer_fail}, where the mouth region in MMI (consistently opened within MMI) has a very different shape compared to the other two datasets. Since in the third fold, IERN was trained with \textit{anger} images from only CK+ and Oulu-CASIA, it failed to recognize \textit{anger} images in MMI.
IERN was rather intended to solve biases caused by dataset characteristics \emph{that are consistently shared across the whole dataset}, \eg, images in MMI have a blue background, CK+ contains mainly gray images, images in Oulu-CASIA are blurry.

The manifestation differences also make the Disentanglement perform much worse than Baseline and IERN in the \textit{anger} expression as shown in Tab.~\ref{tab:fer_results}. On the first and second folds, \textit{anger} faces in the training set are mixed with images from MMI and another dataset. Under such settings, Feature Disentanglement failed to disentangle emotion and context features due to the inconsistent manifestation, resulting in wrong emotion features. Unlike Disentanglement that predicted emotion labels purely based on the extracted emotion features, our IERN still combined emotion with context features in the following steps before applying classiﬁcation, allowing it to overcome the ﬂaws to some extend.

\paragraph{Re-sampling is essentially intervention.}
The most widely adopted method in handling dataset bias is to balance the training data through re-sampling. However, in our o.o.d.\ test setting, the training data does not contain \emph{any} image that shares the same attribute as those in the test set, which makes it \emph{impossible} to directly use re-sampling. 

In fact, re-sampling can be considered as a vanilla form of intervention, since it also aims to solve the bias through balancing the data across all strata of the confounder. For better comparison between IERN and re-sampling, 
we next \emph{modified the setting} by moving 10\% of images (sequence-independent) from the test set into the training set for each attribute in each fold, so that re-sampling can be applied. The performances are reported in Tab.~\ref{tab:fer_results_updated}.

Compared to Baseline, re-sampling achieved a higher performance with 3.04\% improvement. The reason is that re-sampling is approximately intervention, so it can help alleviate the bias and improve the robustness of the model. However, IERN significantly outperformed re-sampling by 10.12\%. This is because by modeling unobserved data using confounder features, IERN can improve not only the balance but also the diversity of training data, while re-sampling does not introduce new data.

\paragraph{Comparison to NWGM-based method.} 
We also compared IERN to the NWGM-based method~\cite{dong2020conta}, given that both methods are based on backdoor adjustment. The confounder features in NWGM are built as a predefined dictionary, by grouping images in the training set according to confounder labels and averaging the extracted features of each group~\cite{dong2020conta, yang2020deconfounded}.

\begin{table*}[t!]
\centering
\caption{Cross-dataset validation for image emotion recognition. Values in cells are the results of Curriculum Learning~\cite{panda2018contemplating}, and values inside brackets are the improvements of IERN. ``Self Test'' means the self-validation (data from the same distribution) experiments, and ``Mean Others'' means the average of cross-datasets (data from different distributions) validations} 
\begin{tabular}{@{}lccccc@{}}
\toprule
\multicolumn{1}{l}{\diagbox[width=9em]{Train on:}{Test on:}} & Deep Sentiment    & Emotion-6         & WEBEmo            & Self Test         & Mean Others       \\ \midrule
Deep Sentiment & 78.74 {[}+ 3.01{]} & 49.76 {[}+ 8.18{]} & 47.79 {[}+ 3.41{]} & 78.74 {[}+ 3.01{]} & 48.78 {[}+ 5.79{]} \\
Emotion-6      & 54.33 {[}+ 5.19{]} & 77.72 {[}+ 1.85{]} & 64.30 {[}+ 0.99{]} & 77.72 {[}+ 1.85{]} & 59.32 {[}+ 3.09{]} \\
WEBEmo         & 68.50 {[}+ 2.13{]} & 78.38 {[}+ 0.53{]} & 81.41 {[}+ 0.31{]} & 81.41 {[}+ 0.31{]} & 73.44 {[}+ 1.33{]} \\ \bottomrule
\end{tabular}
\label{tab:ier_results}
\end{table*}

Using the same \emph{modified setting} as the re-sampling experiment, the results of the NWGM method are shown in Tab.~\ref{tab:fer_results_updated}. Although both re-sampling and NWGM are approximate interventions, NWGM outperformed re-sampling by 2.04\%, suggesting that it is better to model confounders explicitly in the network. Besides, IERN outperformed NWGM by 8.08\%. This is because IERN conducts real intervention by modeling confounder features as trainable parameters, while the pre-computed confounders in NWGM might not be sufficiently representative (see Section~\ref{sec:app_nwgm} for more details).

\subsection{Cross-Dataset Experiments for IER}\label{exp:ier}

\paragraph{Dataset settings.}
Experiments related to image emotion were conducted on three in-the-wild benchmarks, including Deep Sentiment~\cite{you2015robust}, Emotion-6~\cite{panda2018contemplating} and WEBEmo~\cite{panda2018contemplating}. All datasets are based on images collected from the internet, \eg, Flickr, Google. Deep Sentiment contains 1269 images and Emotion-6 has 8350 images, both of which were manually labeled by several human subjects. WEBEmo has around 268,000 images, which were automatically labelled by query keywords. Following \cite{panda2018contemplating}, we report results based on the binary emotion labels, \ie, \textit{positive} and \textit{negative}.

\begin{figure}[t!]
    \centering
    \includegraphics[width=0.23\textwidth]{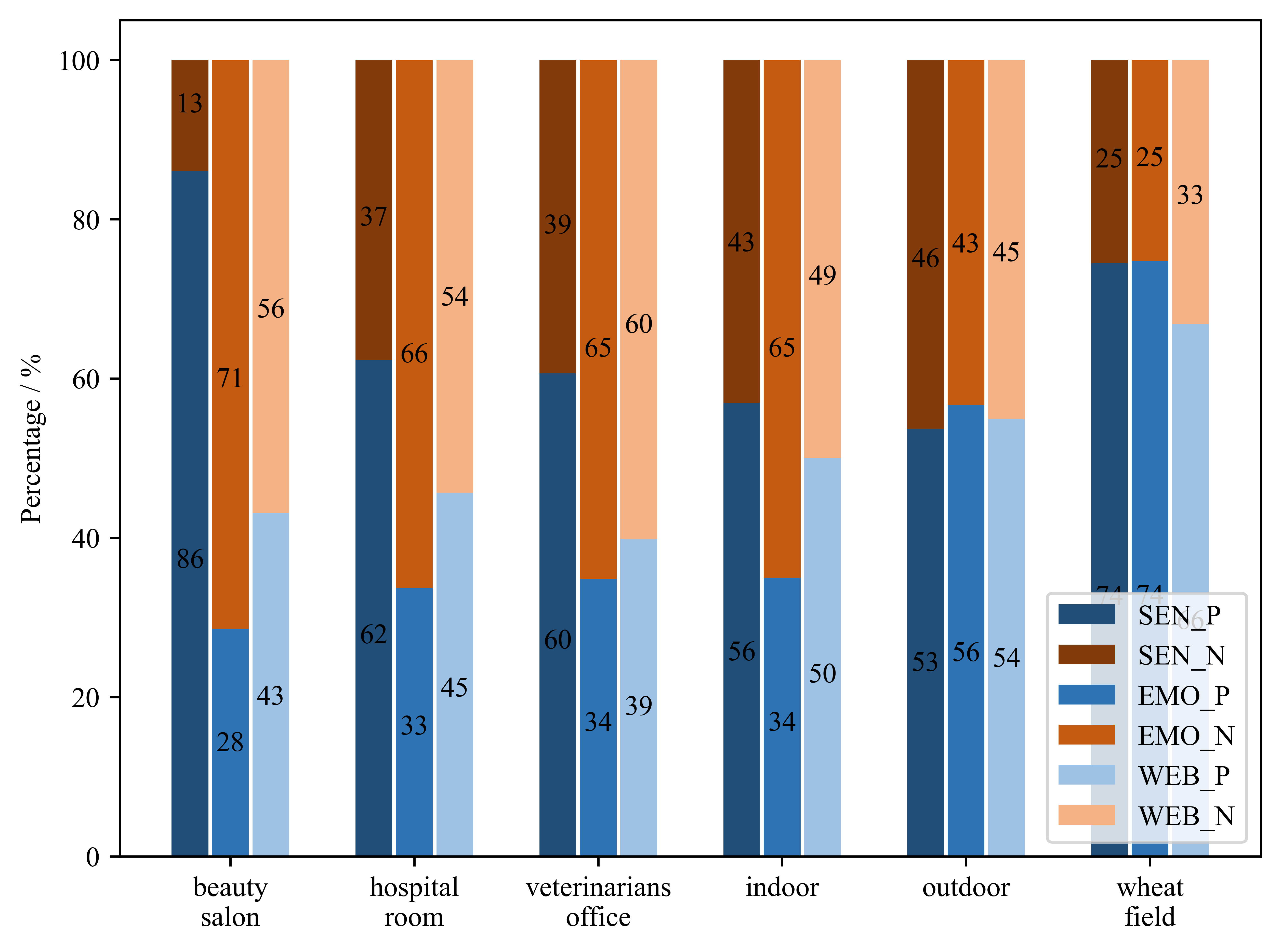}
    \includegraphics[width=0.23\textwidth]{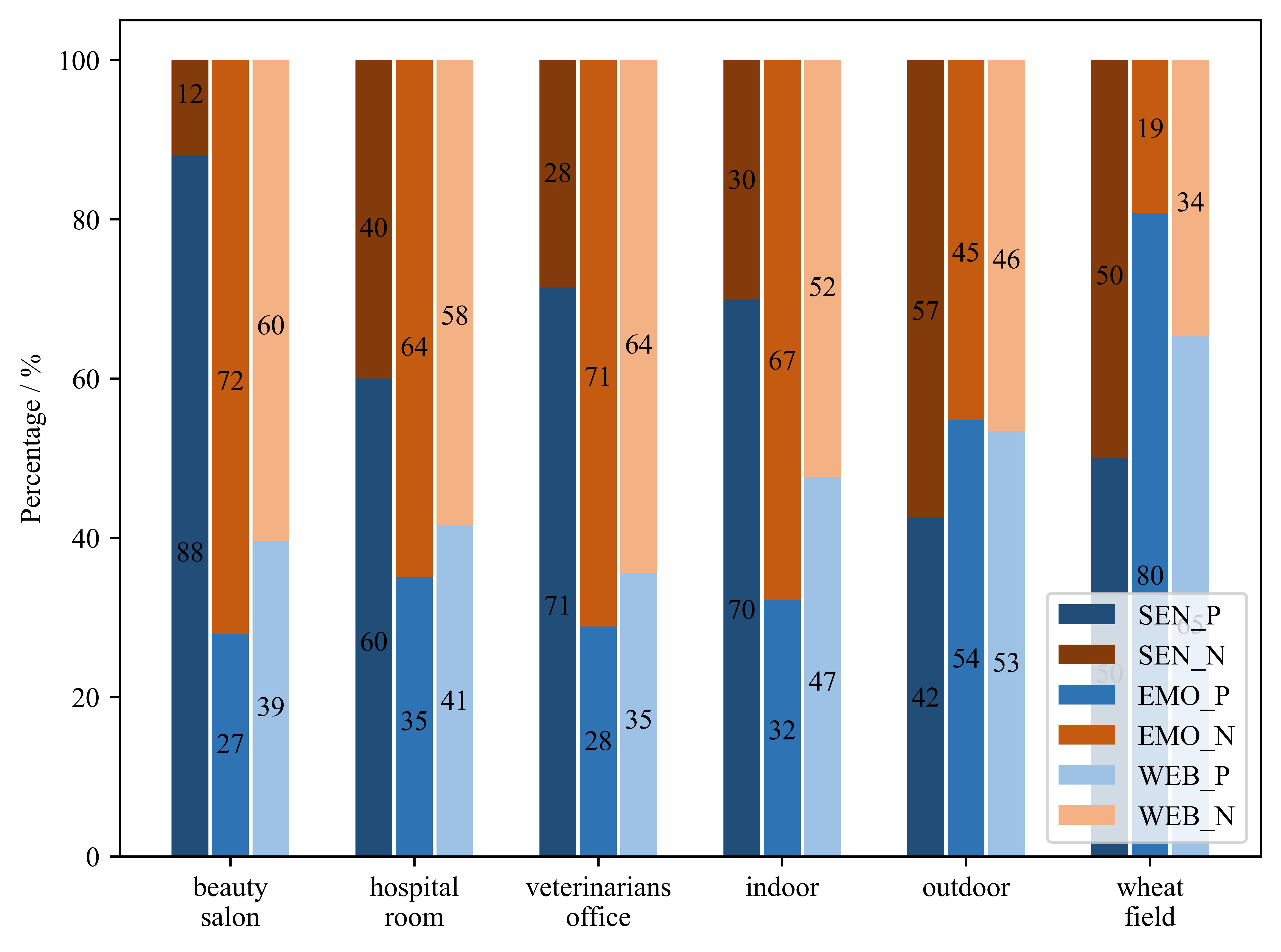} 
    \caption{Emotion distribution within each scene stratum for both training (left) and test sets (right). SEN, EMO and WEB are Deep Sentiment, Emotion-6 and WEBEmo dataset, respectively. P denotes Positive emotion and N denotes Negative emotion (best viewed in high-resolution color).
    }
    \label{fig:ier_distri}
\end{figure}

We followed the same configurations as \cite{panda2018contemplating}, which mainly used 80\% of images for training and the rest for testing. According to the analysis given in \cite{panda2018contemplating}, the \emph{image scene} is the cause of the bias, and is thus the confounder. 
Places365-CNNs~\cite{zhou2017places} was used to detect scene categories, which contains over 300 classes.
To improve error tolerance and reduce computation, we further developed a \emph{two-step approach} to build confounder labels based on the predicted scene labels.
Specifically, scene classes were first grouped into around 30 clusters using k-means, based on their semantic vectors given by GloVe~\cite{pennington2014glove}.Secondly, within each dataset, the clusters were sorted by their importance, which is defined as the product of the class prior and the conditional entropy of emotion, \ie, $I(c) = P(c) \cdot \sum_{e \in \{ p, n \}} P(e| c) \log P(e|c)$, where $c$ refers to the cluster label, and $e$ denotes the binary emotion label, with $p$ being \textit{positive} and $n$ being \textit{negative}. 
The underlying assumption is that if a cluster has more images and more biases, it should be considered more important. 
The 8 most important clusters were selected as the confounder strata, while other clusters were grouped into {\textit{indoor} or \textit{outdoor}}, 
resulting in 10 strata of the confounder for each dataset.

\paragraph{Comparison with state-of-the-art on cross-datasets.}
In order to showcase how dataset bias is alleviated, \cite{panda2018contemplating} designed a cross-datasets validation setting. Following their setting, we also trained IERN on the training set of one dataset to convergence, and directly tested it on test sets of all three datasets. Tab.~\ref{tab:ier_results} shows the results on all datasets, where the numbers in the cells are the accuracy results of the Curriculum Learning method \cite{panda2018contemplating}, while the numbers inside the brackets are the improvements achieved by IERN.

Compared with Curriculum Learning, IERN saw improvements in all settings. The reason is this: while Curriculum Learning aims to solve dataset bias through multi-stage fine tuning, IERN explicitly identifies and models the confounders that caused the bias. This allows intervention to be directly applied to balance emotion data across the confounders, which leads to better performance. 

We noticed that IERN had a lower improvement when trained on WEBEmo compared to training on the other two datasets, which is mainly because WEBEmo is already well-balanced. WEBEmo was proposed in \cite{panda2018contemplating} as a solution to address the bias problem, with emotion data fairly distributed across scenes. To verify, we plot the emotion distribution within each scene stratum for both training and test sets of all datasets. As shown in Fig.~\ref{fig:ier_distri}, compared with Deep Sentiment and Emotion-6, WEBEmo has better-balanced emotion distributions in many scenes. 

We also found that IERN gained less improvement on the self-validation experiments than on cross-dataset ones, especially on Deep Sentiment and Emotion-6. The reason is that self-validation (data from the same distribution) might not be affected by bias~\cite{scholkopf2021toward}. As shown in Fig.~\ref{fig:ier_distri}, within the same dataset, the training and test set have similar bias in the distribution. Thus, although Curriculum Learning might learn biased features during training, it may still achieve good performance on a test set that has the same biases. 

Conversely, in the cases of cross-datasets (data from different distributions) validation, this is no longer true. In particular for many scenes, the biases in Deep Sentiment and Emotion-6 are in opposite directions, \eg, in \textit{beauty\_salon}, the emotion of the former tends to be \textit{positive}, while for the latter it tends to be \textit{negative}. So without proper handling, Curriculum Learning may suffer from opposing biases when trained on one dataset and tested on another, while IERN can remove the bias in training via re-balancing, thus achieving higher performance.

\section{Conclusion}\label{conclusion}
We have studied the dataset bias problem in visual emotion recognition from a causality perspective. 
We have proposed a novel end-to-end trainable framework, IERN, to model the backdoor adjustment theorem of causal intervention in order to achieve unbiased emotion recognition. Extensive experiments on two visual emotion recognition tasks have shown the effectiveness of IERN and its individual components. 
Future works include applying our solution to other vision tasks and considering more complex confounders, \eg, use gender / ethnicity as confounders, automatically learn the number of confounders by embedding clustering algorithms.

\begin{acks}
This study is supported under the RIE2020 Industry Alignment Fund – Industry Collaboration Projects (IAF-ICP) Funding Initiative, as well as cash and in-kind contribution from Singapore Telecommunications Limited (Singtel), through Singtel Cognitive and Artificial Intelligence Lab for Enterprises (SCALE@NTU). This research is also partially supported by FIT Start-up Grant.
\end{acks}

\bibliographystyle{ACM-Reference-Format}
\bibliography{egbib}

\clearpage

\appendix

\section{Technical Details of IERN}

\subsection{Network Architecture}\label{sec:app_network} 

Table~\ref{tab:iern_archi} provides the detailed structure of our implementation of the Interventional Emotion Recognition Network (IERN). Function name from the PyTorch `torch.nn' package is used to denote the layer, \eg, `Conv2d' refers to `torch.nn.Conv2d', the 2D convolution layer. Note that although emotion generator $g_e$ (emotion discriminator $d_e$) and context generator $g_c$ (context discriminator $d_c$) have the same structure, they are declared separately, and their trainable parameters are \textit{not} shared.

\begin{table*}[t!]
\centering
\caption{Detailed structure of all components of IERN. IC: the number of input channels, OC: the number of output channels, K: kernel size, S: stride size, P: padding size, H: height, W: width, $N_e$: the number of emotion classes, $N_c$: the number of confounder strata. Note that $g_e$ (or $d_e$) and $g_c$ (or $d_c$) are declared separately}
\begin{tabular}{@{}ll@{}}
\toprule
\textbf{Network Component}          & \textbf{Detailed Structure}                                                      \\ \midrule 
\multirow{2}{2.5cm}{Backbone $f_b$}   & ResNet-50 (include all but the last average pooling and fully connected layers) \\
                 & Conv2d(IC2048, OC512, K1$\times$1, S1, P0)                                      \\ \bottomrule
\multirow{3}{3.5cm}{Emotion Generator $g_e$ or \\ Context Generator $g_c$}   & Conv2d(IC512, OC512, K3$\times$3, S1, P1)                                       \\
                 & Residual Block: Conv2d(IC512, OC512, K3$\times$3, S1, P1), BatchNorm2d, ReLU             \\
                 & Residual Block: Conv2d(IC512, OC512, K3$\times$3, S1, P1), BatchNorm2d, ReLU             \\ \bottomrule
\multirow{5}{4.4cm}{Emotion Discriminator $d_e$ or \\ Context Discriminator $d_c$}  & Conv2d(IC512, OC512, K4$\times$4, S2, P1), LeakyReLU                            \\
                 & Conv2d(IC512, OC256, K1$\times$1, S1, P0), LeakyReLU                            \\
                 & Conv2d(IC256, OC128, K1$\times$1, S1, P0), LeakyReLU                            \\
                 & Conv2d(IC128, OC64, K1$\times$1, S1, P0), LeakyReLU                             \\
                 & Conv2d(IC64, OC{[}$N_e / N_c${]}, K3$\times$3, S1, P0)      \\ \bottomrule
\multirow{4}{4cm}{Reconstruction Network $g_r$}   & Conv2d(IC1024, OC512, K3$\times$3, S1, P1)                                      \\
                 & Residual Block: Conv2d(IC512, OC512, K3$\times$3, S1, P1), BatchNorm2d, ReLU             \\
                 & Residual Block: Conv2d(IC512, OC512, K3$\times$3, S1, P1), BatchNorm2d, ReLU             \\
                 & Residual Block: Conv2d(IC512, OC512, K3$\times$3, S1, P1), BatchNorm2d, ReLU             \\ \bottomrule
\multirow{2}{2.5cm}{Classifier $f_c$}  & AdaptiveAvgPool2d(H1, W1)                                                \\
                 & Linear(IC512, OC$N_e$)                                         \\ \bottomrule
\end{tabular}
\label{tab:iern_archi}
\end{table*}

\subsection{Implementation}\label{sec:app_alg}

To make clear how different components are optimized sequentially, Algorithm~\ref{alg:train} is detailed with PyTorch-like pseudo codes, shown as below.

\begin{figure}[h!]
    \centering
    \includegraphics[width=0.475\textwidth]{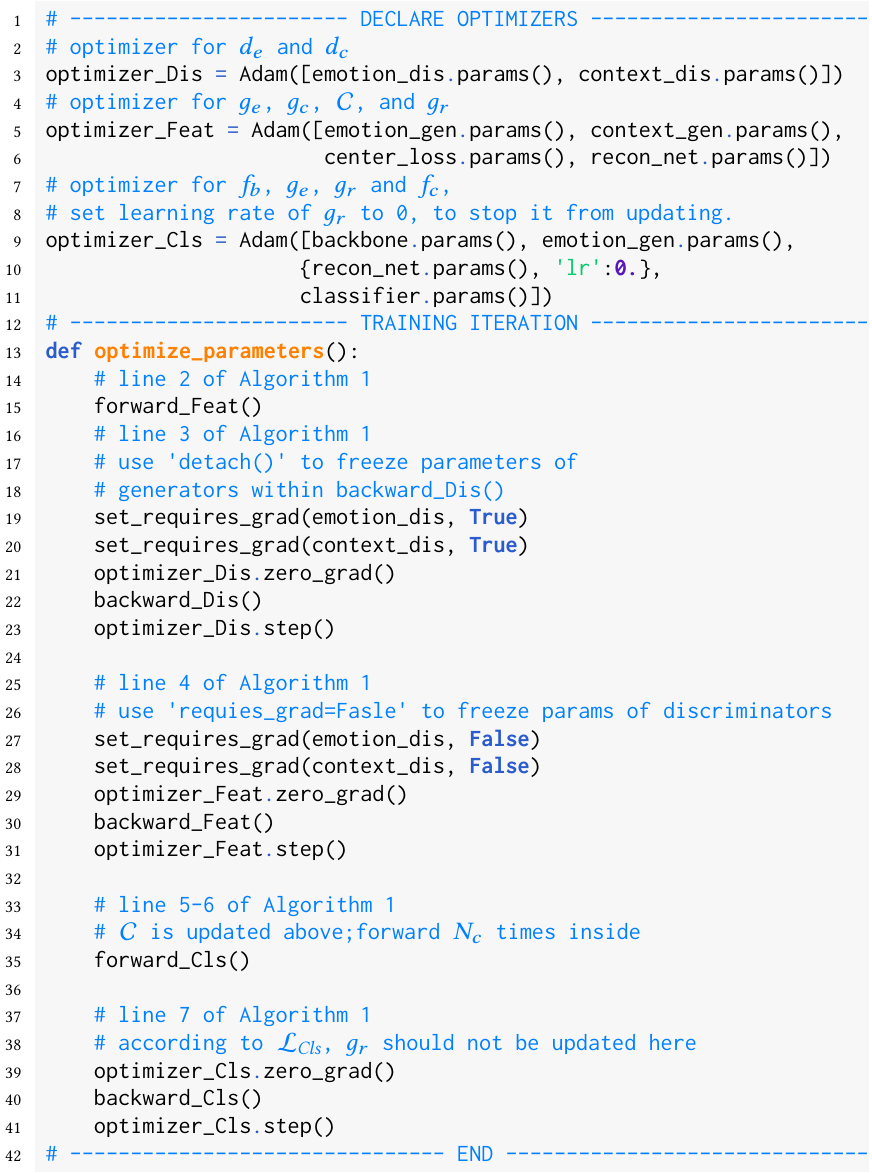} 
\end{figure}

\section{Hyper-parameters searching}\label{sec:app_hyper}

In our experiments, we leveraged an off-the-shelf implementation of center loss\footnote{https://github.com/KaiyangZhou/pytorch-center-loss} for $\mathcal{L}_{\textrm{CB}}$, and found that the magnitude of $\mathcal{L}_{\textrm{CB}}$ is much larger than other loss functions. Therefore, we conducted several experiments on IERN to find the suitable $\lambda_2$ (see Eq.~\eqref{eq:ldo}), by setting $\lambda_1=\lambda_3=1$ and fixing all other training configurations. 

\begin{figure}[t!]
    \centering
    \includegraphics[width=0.47\textwidth]{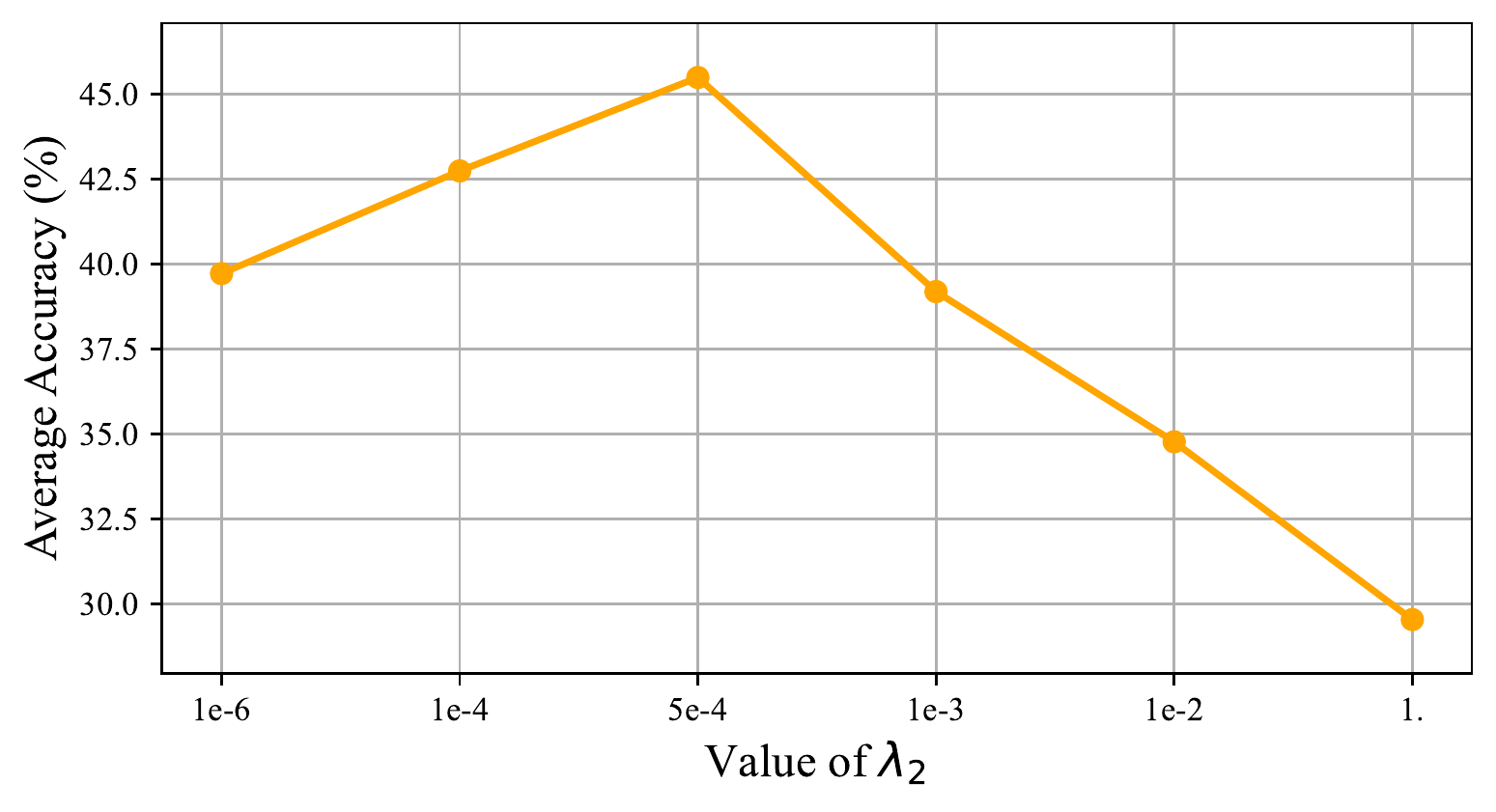} 
    \caption{Hyper-parameters searching on $\lambda_2$ (Eq.~(8) of the main manuscript). We set $\lambda_2=5\times10^{-4}$ in our final model.}
    \label{fig:fer_lambda2}
\end{figure}

As shown in Fig.~\ref{fig:fer_lambda2}, IERN achieved the best performances when $\lambda_2=5\times10^{-4}$, therefore we used this setting in all following experiments. When $\lambda_2$ was set to a smaller value, \eg, $10^{-6}, 10^{-4}$, there is a small drop in the performances. Although confounders were not well learnt in such settings, the final classifier can still be trained sufficiently by $\mathcal{L}_{\textrm{Cls}}$, and the performances decrease were mainly caused by the noisy confounders. However, if $\lambda_2$ was set to a larger value, \eg, $10^{-3}, 10^{-2}, 1$, performances of IERN decreased dramatically. This is because that IERN failed to optimize Feature Disentanglement and Classifier when $\mathcal{L}_{\textrm{CB}}$ was way too large, resulting in an unexpected collapse of the full recognition framework.

\section{Advantages of IERN over NWGM}\label{sec:app_nwgm}

We give a brief introduction of how existing methods~\cite{dong2020conta, yang2020deconfounded} use the Normalized Weighted Geometric Mean (NWGM)~\cite{xu2015show} to approximate the causal intervention.
As detailed in Section~\ref{method:backdoor}, through applying the backdoor adjustment theorem, the model aims to solve
\begin{equation}\label{eq:do}
    P(Y| do(X)) = \sum_{d}^{}P(Y|X, D=d)P(D=d),
\end{equation}
where $X$ and $D$ are the intended feature and confounder feature, respectively. 
Here $P(Y|X, d)$ is normally obtained from the predicted logits of a classification network, \ie, $\sigma(f(X, d))$, where $\sigma$ denotes the softmax function. Thus, Eq.~\eqref{eq:do} can be written as
\begin{equation}
    P(Y| do(X)) :=  \mathbb{E}_d[\sigma(f(X, d))].
\end{equation}

NWGM is then applied to move the outer expectation into the softmax function, that is
\begin{equation}\label{eq:nwgm}
    \mathbb{E}_d[\sigma(f(X, d))] \stackrel{\text{\tiny NWGM}}{\approx} \sigma( \mathbb{E}_d[f(X, d)] ).
\end{equation}

Supposed that the operation of combining $X$ and $D$ is linear, then Eq.~\eqref{eq:nwgm} can be further deduced as
\begin{equation} 
\begin{split}
\mathbb{E}_d[f(X, d)] & = \mathbb{E}_d[W_1 x + W_{2} \cdot g(d)] \\
 & = W_1 x + W_{2} \cdot \mathbb{E}_d[g(d)],
\end{split}
\end{equation}
where $W_1$ and $W_2$ both refer to trainable parameters of the fully connected layers. And as a general approach~\cite{dong2020conta, yang2020deconfounded}, $\mathbb{E}_d[g(d)]$ is modelled as a attention layer, where the query is set to the base feature and the key is set to a predefined confounder dictionary. 

Although existing works have demonstrated the effectiveness of the NWGM-based method, at least three shortcomings can be found. Firstly, as pointed out in Section~\ref{exp:fer}, the confounder features are not sufficiently representative when constructed via a predefined / pre-computed manner. Secondly, it may not be correct to assume that $X$ and $D$ are linearly combined, given that $D$ could be very complex in practice. Thirdly, NWGM is essentially an approximation and may introduce unintended noise.

By efficiently embedding the backdoor adjustment theorem, our proposed IERN overcomes the aforementioned limitations. Firstly, IERN directly learns confounder features through disentanglement and leveraging the center loss function. Secondly, IERN reuses the reconstruction network to combine $X$ and $D$, removing the restrictive assumption of the linear combination operation. Thirdly, IERN does not involve NWGM approximation, instead, it forwards the combined features $N_c$ times to \emph{do} the \emph{real} intervention. Comparison experiments in Section~\ref{exp:fer} verify the superiority of IERN over NWGM-based methods.

\end{document}